\documentclass{Interspeech}

\interspeechcameraready

\title{Articulatory clarity and variability before and after surgery for tongue cancer{}}

\author[affiliation={1,2}]{Thomas}{Tienkamp}
\author[affiliation={1}]{Fleur}{van Ast}
\author[affiliation={1}]{Roos}{van der Veen}
\author[affiliation={3,4}]{Teja}{Rebernik}
\author[affiliation={1}]{Raoul}{Buurke}
\author[affiliation={1}]{Nikki}{Hoekzema}
\author[affiliation={1}]{Katharina}{Polsterer}
\author[affiliation={1}]{Hedwig}{Sekeres}
\author[affiliation={5,6}]{Rob}{van Son}
\author[affiliation={1}]{Martijn}{Wieling}
\author[affiliation={2}]{Max}{Witjes}
\author[affiliation={2}]{Sebastiaan}{de Visscher}
\author[affiliation={1}]{Defne}{Abur}

\affiliation{CLCG}{University of Groningen}{the Netherlands}
\affiliation{}{University Medical Center Groningen}{the Netherlands}
\affiliation{LPP}{CNRS/Sorbonne Nouvelle}{France}
\affiliation{BCLS}{Vrije Universiteit Brussel}{Belgium}
\affiliation{}{University of Amsterdam}{the Netherlands}
\affiliation{}{Netherlands Cancer Institute}{the Netherlands}
\email{t.b.tienkamp@rug.nl, d.abur@rug.nl}

\usepackage{comment}

\begin{document}

\maketitle

\begin{abstract}
    
        Surgical treatment for tongue cancer can negatively affect the mobility and musculature of the tongue, which can influence articulatory clarity and variability. In this study, we investigated articulatory clarity through the vowel articulation index (VAI) and variability through vowel formant dispersion (VFD). Using a sentence reading task, we assessed 11 individuals pre and six months post tongue cancer surgery, alongside 11 sex- and age-matched typical speakers. Our results show that while the VAI was significantly smaller post-surgery compared to pre-surgery, there was no significant difference between patients and typical speakers at either time point. Post-surgery, speakers had higher VFD values for /\textipa{i}/ compared to pre-surgery and typical speakers, signalling higher variability. Taken together, our results suggest that while articulatory clarity remained within typical ranges following surgery for tongue cancer for the speakers in our study, articulatory variability increased.

\end{abstract}

\section{Introduction}
Surgical treatment for tongue squamous cell carcinoma (TSCC) results in anatomical alterations which may complicate speech articulation \cite{kreeft_speech_2009}. Scar tissue, or tethering of the residual tongue to the floor of the mouth may reduce tongue mobility \cite{tienkamp_quantifying_2024}. In addition, surgery induces changes to the tongue musculature, which may lead to reduced control over the residual tongue. One way to assess surgery-induced speech changes is by analysing spectrotemporal features from acoustic recordings. Given the relationship between specific acoustic features and tongue movements, acoustic measures can provide information regarding the articulatory basis of the observed changes \cite{mefferd_articulatory--acoustic_2010}. Yet, acoustic methods are not commonly employed to assess speech function following surgery for TSCC as they are more time-intensive compared to perceptual evaluations \cite{dwivedi_evaluation_2009}.

When used, acoustic assessment of speech in individuals with speech difficulties, including those following surgery for TSCC, often focuses on vowel formant frequency metrics. Metrics such as the vowel space area (VSA) capitalise on the fact that the first and second formant (\textit{F}\textsubscript{1} and \textit{F}\textsubscript{2}) have been linked to specific tongue movements. Where tongue and jaw height primarily modify \textit{F}\textsubscript{1}, tongue frontedness is mostly associated with \textit{F}\textsubscript{2} \cite{mefferd_articulatory--acoustic_2010}. Thus, larger underlying tongue movements correspond to a larger VSA, potentially enhancing intelligibility by improving phoneme differentiation. Indeed, a larger VSA has been shown to be associated with better speech intelligibility in speakers treated for TSCC \cite{de_bruijn_objective_2009}.

In general, acoustic studies have shown that the VSA is reduced following surgery for TSSC compared to pre-surgery \cite{guo_speech_2023, takatsu_phonologic_2017} or typical speakers \cite{whitehill_acoustic_2006, de_bruijn_objective_2009}, though not all studies have found significant changes in the VSA post-surgery \cite{tienkamp_objective_2023, laaksonen_speech_2010}. Tumour size, which ranges from T1 (smallest) to T4 (largest), influences post-surgery speech function as individuals with larger tumours (T3-T4) show greater reductions in the VSA post-surgery than those treated for smaller tumours (T1-T2) \cite{guo_speech_2023, takatsu_phonologic_2017}. The tumour location further influences post-surgery speech function, with more typical speech following surgery for more posterior tumours compared to anterior tumours \cite{hagedorn_characterizing_2014}. Moreover, Hagedorn and colleagues \cite{hagedorn_characterizing_2014} reported that while \textit{F}\textsubscript{1}-range is mostly preserved, large reductions are found in \textit{F}\textsubscript{2}, indicating problems with tongue fronting. Compensatory strategies to maintain vowel distinctions can emerge following TSCC surgery, where shifts in one vowel may result in changes to another to preserve contrast. For example, Laaksonen and colleagues found that while individual formants changed significantly post-surgery, the overall VSA remained stable \cite{laaksonen_speech_2010}.

Formant metrics, such as the VSA, quantify the size of the acoustic working space but do not capture finer aspects of articulatory control, such as articulatory variability. Articulatory variability reflects the stability with which speech sounds are produced. While variability is inherent in speech due to the complex coordination of subsystems (i.e., the articulatory, laryngeal, and respiratory systems), excessive variability has been linked to less efficient and more error-prone speech patterns \cite{abbiati_speech_2023}. Although articulatory variability has not been formally studied in individuals with TSCC, findings by Whitehill and colleagues \cite{whitehill_acoustic_2006} suggest that speakers with a partial glossectomy exhibit greater speech variability, as evidenced by higher standard deviations in vowel formant frequencies compared to typical speakers. Moreover, Grimm and colleagues \cite{grimm_effects_2017} hypothesise that tongue tip control may be reduced as speakers prefer /s/ configurations that require less tongue tip elevation and fronting post-surgery compared to typical speakers. Assessing articulatory variability alongside articulatory clarity may provide a more comprehensive understanding of post-surgical speech outcomes, which in turn may facilitate the development of more targeted speech interventions.

Therefore, the purpose of the current study was twofold. Our first aim was to evaluate articulatory clarity as quantified by vowel acoustics in individuals before and after surgical treatment for TSCC and in typical speakers. Specifically, we assessed articulatory clarity through the vowel articulation index (VAI). We used the VAI because it is less susceptible to outliers than the VSA, making it more sensitive to articulatory changes \cite{skodda_vowel_2011}. We did not predict a reduced VAI post-surgery compared to pre-surgery and typical speakers as our speakers were mostly treated for small tumours. Our second aim was to quantify articulatory variability to assess the stability of speech post-surgery for TSCC. Articulatory variability was assessed through vowel formant dispersion (VFD; \cite{karlsson_vowel_2012}), which quantifies token-to-token variation by computing the Euclidean distance in formant-space between a produced vowel and its respective centre. If surgery negatively affects tongue control due to changes to its musculature, we predict higher levels of variability post-surgery compared to pre-surgery and typical speakers. Conversely, scar-tissue could result in more rigid movement patterns, which could result in reduced levels of variability compared to pre-surgery and typical speakers.

\section{Method}
\subsection{Speakers} 
The present study is part of a larger project approved by the institution's Medical Ethics Review Board (NL79242.042.21). All participants provided written informed consent before their participation. Twelve individuals scheduled for TSCC surgery between November 2022 and March 2024 were recruited, with data collected pre- and six months post-surgery. One speaker (OC10) was unavailable for follow-up, leaving eleven speakers (7 males, 4 females; mean age = 62 years, range = 40–77) for analysis. Two participants (OC02 \& OC04) were treated for a tumour on the posterior $1/3$ of the tongue whereas the other nine for a tumour on the anterior $2/3$ of the tongue. For most speakers, the tumour was lateral; for one speaker (OC07), the tumour was on the tongue's midline. The tongue was locally closed for most speakers, but two speakers (OC01 \& OC02) received reconstruction by means of a radial forearm free flap (RFFF). Individuals received treatment for T1 tumours (n = 7), T2 tumours (n = 3), T3 tumours (n = 1), or a carcinoma in situ (Tis, n = 1). The individual with the carcinoma in situ was clinically staged at cT2, but pathologically revised to Tis. Demographic and clinical information is summarised in Table \ref{tab:clinical_details}. In addition, eleven typical speakers who were sex- and age-matched to the TSCC group also participated (7 males; 4 females; mean age = 62 years; age range = 39-77). All speakers were native speakers of Dutch, and denied a history of speech, language, or neurological impairments.

\subsection{Procedures}
Speakers were recorded in the sound booth of the mobile laboratory SPRAAKLAB \cite{rebernik_spraaklab_2025} with an omni-directional microphone (Shure MX-153T) angled 45\textdegree~from the mouth with a seven centimetre mic-to-mouth distance. Speech was recorded at a 22,050 Hz sampling rate and digitised using a Focusrite Scarlett Solo (2nd gen) soundcard. As part of a larger experimental protocol, speakers produced three repetitions of five phonemically balanced Dutch sentences \cite{luts_development_2014}. From these sentences, we manually segmented at least twelve tokens each of the following vowels per speaker and time point in a stressed position: /\textipa{i}, \textipa{e:}, \textipa{E}, \textipa{a:}, \textipa{o:}, \textipa{u}/. This resulted in a total of 3,010 vowels with approximately 90 vowels per speaker (standard deviation (SD) = 2.7).

\begin{table}[h]
\caption{Demographical and clinical information. F = female, M = male, Tis = carcinoma in situ, A = anterior $2/3$ of the tongue, P = posterior $1/3$ of the tongue, FOM = floor of the mouth.}
  \label{tab:clinical_details}
  \centering
\begin{tabular}{ccccc}
\toprule
\textbf{Speaker} & \textbf{Sex} & \textbf{Age} & \textbf{T-stage} & \textbf{Location} \\
\midrule
OC01             & F            & 75.2         & T3               & A                 \\
OC02             & M            & 41.5         & T2               & P + FOM           \\
OC03             & M            & 54.8         & T1               & A                 \\
OC04             & F            & 77.3         & T1               & P                 \\
OC05             & M            & 55.5         & T1               & A                 \\
OC06             & M            & 68.5         & T2               & A                 \\
OC07             & F            & 61.6         & T1               & A                 \\
OC08             & M            & 62.1         & T1               & A                 \\
OC09             & F            & 40.1         & T1               & A                 \\
OC11             & M            & 76.3         & Tis (cT2)        & A                 \\
OC12             & M            & 71.9         & T2               & A                 \\
\bottomrule
\end{tabular}
\end{table}

\subsection{Acoustic analysis and variable construction}
Vowel formants (\textit{F}\textsubscript{1} and \textit{F}\textsubscript{2}) were extracted using a Praat script in Praat 6.4 \cite{boersma_praat_2024} using the Burg algorithm within the middle 25\% of the vowel, with time steps of 0.01 seconds and an analysis window of 0.025 seconds. Given that the accuracy of formant tracking is both speaker- and vowel-dependent \cite{escudero_cross-dialect_2009}, we extracted all vowels using formant ceilings of 4,000-6,000 Hz with 500 Hz steps. Following Escudero and colleagues \cite{escudero_cross-dialect_2009}, the `optimal' formant ceiling was the one that yielded the lowest variation per vowel. Any remaining formant mistrackings were manually removed through visual inspection (n = 50, 1.7\%), leaving 2,960 vowels for analysis.

To assess articulatory function, we calculated the vowel articulation index \cite{skodda_vowel_2011} for each speaker at each time point. The VAI was calculated with formant frequencies in Hertz \cite{skodda_vowel_2011}.
To assess articulatory variability, we used the vowel formant dispersion metric \cite{karlsson_vowel_2012}. The VFD is normally computed as the Euclidean distance in formant-space from a given vowel token to the centre of a speaker's vowel space (i.e.,~the mean \textit{F}\textsubscript{1}-\textit{F}\textsubscript{2} of all vowels). Instead of using the overall vowel space midpoint, we calculated the Euclidean distance between each vowel token and its vowel-specific center to more directly assess within-phoneme variability (e.g.,~the median \textit{F}\textsubscript{1}-\textit{F}\textsubscript{2} of /\textipa{i}/). We opted for the median as it is more robust to outliers than the mean. Following recommendations by Adank and colleagues \cite{adank_comparison_2004}, formant frequencies were \textit{z}-transformed (i.e.,~Lobanov normalised) to eliminate anatomical differences in order to make formants comparable across speakers.


\begin{figure*}[t]
\begin{center}
\includegraphics[width=17cm]{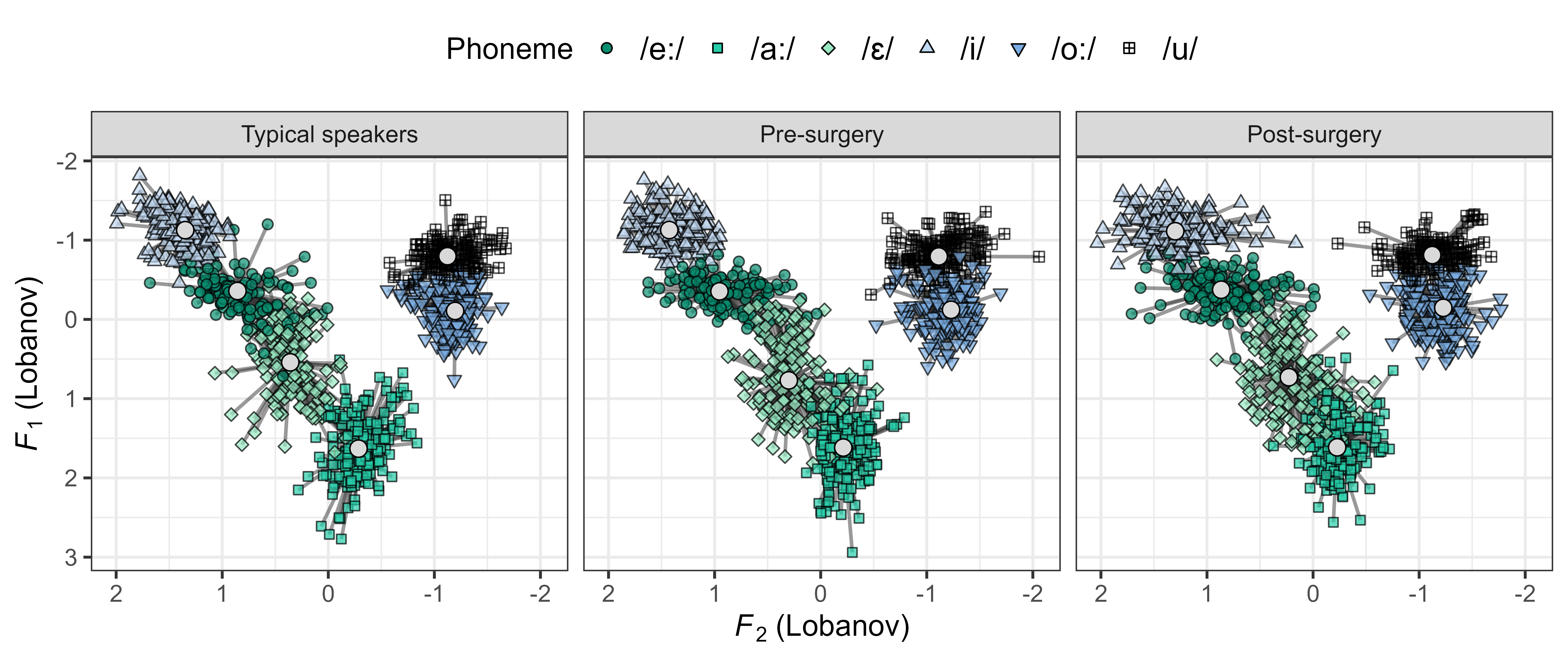}
\caption{\textit{F}\textsubscript{1} and \textit{F}\textsubscript{2} of vowels on a Lobanov-scale per group and time. Points represent vowel tokens and are connected by a dark grey line to the median \textit{F}\textsubscript{1}-\textit{F}\textsubscript{2} coordinate of each phoneme, shown as a large light grey point. Points are coloured and shaped by phoneme.}
\label{fig:VFD_descriptive}
\end{center}
\end{figure*}

\subsection{Statistical analysis}
The VAI and VFD data were statistically analysed in R version 4.4.2 \cite{r_core_team_r_2024} using linear mixed-effects regression \cite{bates_fitting_2015}. For articulatory clarity, our hypothesis model fitted the VAI as a function of subgroup (typical, pre-surgery, post-surgery), with a by-speaker random intercept. For articulatory variability, our hypothesis-testing model fitted the VFD as a function of subgroup, with by-speaker and by-word random intercepts. We exploratorily assessed whether the addition of phoneme (/\textipa{i}, \textipa{e:}, \textipa{E}, \textipa{a:}, \textipa{o:}, \textipa{u}/) and the interaction between subgroup and phoneme improved the fit of the model. We also assessed the influence of sex and age in an exploratory manner for both models. All numerical variables were centred around the mean and the $\alpha$-level was set at 0.05. We concluded our analysis by checking model assumptions and employing model criticism. That is, we checked whether outliers drove the absence or presence of statistically significant effects by removing data points with an absolute residual exceeding 2.5 SD from their fitted value. We only used this trimmed dataset when outliers drove the absence or presence of statistically significant effects \cite{baayen_analyzing_2008}. Pairwise comparisons were made with the \textit{emmeans} package \cite{lenth_emmeans_2022} and corrected using the false discovery rate procedure \cite{benjamini_controlling_1995}.

\section{Results}
\begin{figure*}[t]
\begin{center}
\includegraphics[width=17cm]{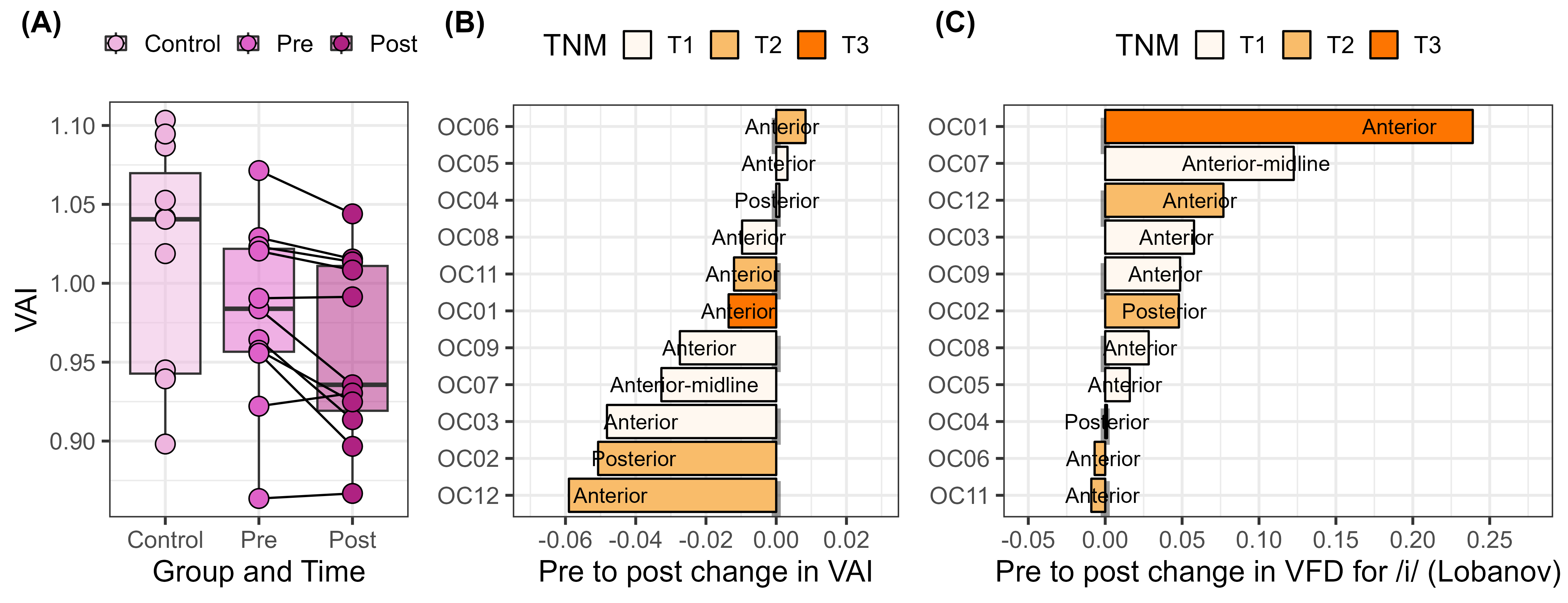}
\caption{(A) VAI per group and time point. Pre- and post-surgery data points from the same individual are connected by a black line. (B) Pre-surgery to post-surgery change in VAI. A negative value indicates a smaller VAI post-surgery compared to pre-surgery and vice-versa. (C) Pre-surgery to post-surgery change in VFD for /\textipa{i}/ on a Lobanov scale. A positive value indicates higher levels of variability post-surgery compared to pre-surgery and vice-versa. In both panels (B) and (C), OC11, who had a carcinoma in situ, is coloured according to the clinical diagnosis (cT2).}
\label{fig:combined}
\end{center}
\end{figure*}

The group-level visualisation of the vowel space and token-to-token variability for typical speakers, speakers pre-, and post-surgery is provided in Figure \ref{fig:VFD_descriptive} and for the VAI in Figure \ref{fig:combined}-A. The VAI results are based on the full dataset. Pairwise comparisons indicated that the VAI was significantly lower post-surgery (mean VAI post-surgery = 0.96, SD = 0.06) compared to pre-surgery (mean VAI pre-surgery = 0.98, SD = 0.06; $\beta$ = -0.02, \textit{T} = -3.1, $\textit{p} = 0.03$). The change in VAI from pre- to post-surgery is plotted per individual in Figure \ref{fig:combined}-B. There was no significant difference between the VAI of typical speakers (mean VAI typical speakers = 1.01, SD = 0.07) and speakers with TSCC pre-surgery ($\beta$ = 0.03, \textit{T} = 1.2, $\textit{p} = 0.23$) or post-surgery ($\beta$ = -0.05, \textit{T} = -2.0, $\textit{p} = 0.08$). The exploratory analysis indicated that neither sex ($\chi^2$(1) = 3.46, $\textit{p} = 0.06$) nor age ($\chi^2$(1) = 0.86, $\textit{p} = 0.35$) significantly improved the model.


The VFD results are based on the trimmed dataset (57 datapoints removed, 1.93\%) in which we log-transformed the VFD as the residuals were not normally distributed. The addition of phoneme ($\chi^2$(5) = 30.6, $\textit{p} < .001$) and interaction between phoneme and subgroup significantly improved the base model ($\chi^2$(10) = 21.4, $\textit{p} = 0.02$). Pairwise comparisons indicated that post-surgery, individuals treated for TSCC showed higher levels of variability for /\textipa{i}/ compared to pre-surgery ($\beta$ = 0.2 log-Lobanov distance, \textit{T} = 2.9, $\textit{p} = 0.005$) and typical speakers ($\beta$ = 0.31 log-Lobanov distance, \textit{T} = 4.0, $\textit{p} < .001$). The change in variability for /\textipa{i}/ from pre- to post-surgery is plotted per individual in Figure \ref{fig:combined}-C. A trend-effect indicated that individuals treated for TSCC showed higher levels of variability for /\textipa{E}/ compared to typical speakers, both pre-surgery ($\beta$ = 0.19 log-Lobanov distance, \textit{T} = 2.3, $\textit{p} = 0.07$) and post-surgery ($\beta$ = 0.17 log-Lobanov distance, \textit{T} = 2.0, $\textit{p} = 0.07$), but no difference between pre- and post-surgery was found ($\beta$ = -0.02 log-Lobanov distance, \textit{T} = -0.3, $\textit{p} = 0.78$). None of the other pairwise comparisons resulted in statistically significant differences (all $\textit{p}\textrm{'s} > 0.05$). The exploratory analysis further indicated that neither sex ($\chi^2$(1) = 0.39, $\textit{p} = 0.53$) nor age ($\chi^2$(1) = 0.12, $\textit{p} = 0.73$) significantly improved the model.

\section{Discussion}
In this study, we assessed the articulatory clarity (Aim 1) and variability (Aim 2) of speakers with TSCC pre- and post-surgery alongside typical speakers. For our first aim, in which we assessed articulatory clarity as quantified by the VAI, we made two predictions. First, we predicted no VAI reduction post-surgery compared to typical speakers as the speakers included in our study were mostly treated for smaller tumours. The results support this prediction, as no differences were found between typical speakers and speakers pre- or post-surgery for TSCC. These findings suggest that individuals with TSCC have typical speech clarity pre-surgery, and good speech outcomes post-surgery when measured acoustically. Our results are in line with previous work that found more typical speech outcomes in individuals treated for smaller tumours as compared to larger tumours \cite{guo_speech_2023, takatsu_phonologic_2017}.

Second, we predicted no reduction of the VAI post-surgery compared to pre-surgery. Contrary to this prediction, our results showed a significant decrease of the VAI post-surgery compared to pre-surgery. However, reductions were small (mean reduction of 0.02, ranging from a -0.06 decrease to a 0.008 increase) and it remains largely unknown for the VAI what the minimal detectable change is (i.e.,~whether the observed change is outside the measurement error), and what a clinically meaningful change is (i.e.,~the amount of change that is perceptually relevant). While the second question remains unexplored, there is an initial indication as to the measurement error. That is, Skodda and colleagues \cite{skodda_impairment_2012} tested 40 typical speakers on average 20 months apart and their raw data suggest a mean difference of 0.01 in the VAI of typical speakers. Using a conservative cut-off of 0.02 (i.e.,~double the mean difference), our results suggest that 5/11 speakers show a reduction in VAI as a result of surgery for TSCC. It seems unlikely that these changes resulted in intelligibility loss as they fall within typical ranges, which was also reflected in the statistically non-significant difference between speaker groups. No clear pattern was found between the change in VAI post-surgery and tumour size and location. For example, two individuals with T2 tumours showed the largest reductions, while the other two with T2 tumours had changes within the measurement error. The lack of a clear pattern may reflect generally small changes or the wide variability in speech outcomes after TSCC treatment \cite{bressmann_speech_2021}.

Our second aim was to assess articulatory variability, quantified by the VFD. We did not form any precise predictions, but preliminary evidence suggested that articulatory variability may increase following surgery for TSCC \cite{whitehill_acoustic_2006, grimm_effects_2017}. Consistent with prior studies, we found increased variability for the high-front vowel /\textipa{i}/ post-surgery compared to pre-surgery and typical speakers, driven strongly by the speaker with the largest tumour (T3) and RFFF reconstruction (OC01). Since this speaker’s VAI remained unchanged, the results suggest that while extreme articulatory targets can still be reached post-surgery, they become less stable (i.e.,~more variable). Reduced tongue control may be due to the flap itself, since it only moves passively as it is not functionally integrated \cite{bressmann_speech_2021}, or due to surgery-induced changes to the tongue musculature. Besides the significant increase in VFD for /\textipa{i}/ and the trend-effect for the mid-front vowel /\textipa{E}/, no other changes were found in articulatory variability, which suggests that variability is only affected for sounds that require potentially affected movements, namely tongue raising and/or fronting \cite{tienkamp_articulatorykinematic_2025}. While there was no clear pattern based on tumour size and location with regards to articulatory variability, it is worth noting that speaker OC07, who was treated for a small tumour on the tongue's anterior midline, showed both a reduction in the VAI as well as the second largest increase in variability. Future studies may investigate whether tumours on the midline result in larger changes as compared to tumours located laterally on the tongue.

A second explanation for the observed increased variability of /\textipa{i}/ could be that compensatory strategies have not fully been established. As speakers are still exploring the motor space of the tongue, jaw, and lips for the configuration that yields the optimal acoustic output, they initially become more variable as this increased variability may be beneficial for motor learning \cite{dhawale2017role}. If this is indeed the case, one would expect that variability would return to pre-surgery levels when strategies have been established. The moment at which strategies are fully established remains relatively unclear, with some studies suggesting that improvement after the 12-month mark is still possible \cite{matsui_factors_2007}.

When interpreting the results of the current work, it is also necessary to consider the study's limitations. First, we only report on a relatively small clinical sample that was primarily treated for T1 tumours (n = 7; 64\%). A larger sample size, as well as a more equal spread between tumour severity are needed in order to corroborate the results reported here. Second, future work should explore to what extent the decrease in VAI or increased variability of /\textipa{i}/ affected speech intelligibility or acceptability since we did not run perceptual tests.

\section{Conclusion}
This longitudinal study assessed articulatory clarity (using VAI) and variability (using VFD) in speakers before and after surgery for TSCC using vowel acoustics. The results indicate no significant difference in VAI between speakers pre- and post-surgery and typical speakers, though a small significant reduction was found post-surgery compared to pre-surgery. Articulatory variability was significantly higher for the high front vowel /\textipa{i}/ post-surgery compared to both pre-surgery and typical speakers, with the highest increase in variability for a speaker who received RFFF reconstruction. It remains to be explored whether this observed increase in variability post-surgery is due to loss of articulatory control or due to motor space exploration to optimise compensatory behaviour. Taken together, our results suggest that even though articulatory clarity remains within typical ranges for individuals who underwent surgery for TSCC, post-surgery articulatory variability was increased compared to pre-surgery and typical speakers.

\newpage
\section{Acknowledgements}
We are grateful to all speakers who participated in this project. We acknowledge funding from the Center for Language and Cognition and from the Research School of Behavioral and Cognitive Neurosciences of the University of Groningen. We further acknowledge funding from Atos Medical (Hörby, Sweden) awarded to the Netherlands Cancer Institute.

\bibliographystyle{IEEEtran}
\bibliography{references}

\end{document}